\documentclass{article}

\PassOptionsToPackage{numbers}{natbib}
\usepackage[preprint]{neurips_2023}

\usepackage[utf8]{inputenc} 
\usepackage[T1]{fontenc}    
\usepackage{url}            
\usepackage{booktabs}       
\usepackage{amsfonts}       
\usepackage{nicefrac}       
\usepackage{microtype}      
\usepackage{xcolor}
\usepackage{url}

\usepackage{amsmath}
\usepackage{amssymb}
\usepackage{amsthm}
\usepackage{placeins}
\usepackage{wrapfig}
\usepackage{subfigure}
\usepackage{graphicx}
\usepackage{nicefrac}
\newtheorem{theorem}{Theorem}

\newcommand{\E}[2]{\operatorname{\mathbb{E}}_{#1}\left[#2\right]}

\newcommand{\density}{p}

\newcommand{\kl}[2]{\mathrm{D_{KL}}\left(#1\;\middle\|\;#2\right)}

\newcommand{\ent}{\mathcal{H}}

\newcommand{\voidarg}{{\,\cdot\,}}

\newcommand{\sspace}{\mathcal{S}}
\newcommand{\aspace}{\mathcal{A}}

\newcommand{\state}{\mathbf{s}}

\newcommand{\st}{{\state_t}}

\newcommand{\stp}{{\state_{t+1}}}

\newcommand{\pdyn}{\density}

\newcommand{\action}{\mathbf{a}}

\newcommand{\at}{{\action_t}}

\newcommand{\atp}{{\action_{t+1}}}

\newcommand{\reward}{r}

\newcommand{\rmin}{r_\mathrm{min}}
\newcommand{\rmax}{r_\mathrm{max}}

\newcommand{\Q}{Q}

\newcommand{\policy}{\pi}

\newcommand{\gauss}{\mathcal{N}}

\newcommand{\reals}{\mathbb{R}}

\newcommand{\discount}{\gamma}

\usepackage{algorithm}
\usepackage{algorithmicx}  
\usepackage{algpseudocode}

\title{Careful at Estimation and Bold at Exploration}

\author{
  $\text{\textbf{Xing Chen}}^{1}$~~~~~~~~~~
  $\text{\textbf{Yijun Liu}}^{1}$~~~~~~~~~~
  $\text{\textbf{Zhaogeng Liu}}^{1}$~~~~~~~~~~
  $\text{\textbf{Hechang Chen}}^{1}$\\
  $\text{\textbf{Hengshuai Yao}}^{2}$~~~~~~~~~~
  $\text{\textbf{Yi Chang}}^{1}$\\
  ${}^{1}$School of Artificial Intelligence, Jilin University \\
  ${}^{2}$Department of Computing Science, University of Alberta\\
}

\begin{document}

\maketitle

\begin{abstract}
Exploration strategies in continuous action spaces are often heuristic due to the infinite actions. However, these kinds of methods cannot derive a general conclusion. In prior work, it has been shown that deterministic policy reinforcement learning (DPRL) is better for handling continuous action RL tasks, and exploration guided by policy gradient is beneficial. However, DPRL has two prominent issues: aimless exploration and policy divergence, and the policy gradient for exploration is only sometimes helpful due to inaccurate estimation. Based on the double-Q function framework, we introduce a novel exploration strategy to mitigate these issues, separate from the policy gradient. We first propose the greedy Q softmax update schema for Q value update. The expected Q value is derived by weighted summing the conservative Q value over actions, and the weight is the corresponding greedy Q value. Greedy Q takes the maximum value of the two Q functions, and conservative Q takes the minimum value of the two different Q functions. For practicality, this theoretical basis is then extended to allow us to combine action exploration with the Q value update, except for the premise that we have a surrogate policy that behaves like this exploration policy. In practice,  we construct such exploration policy with a few sampled actions, and to meet the premise, we learn such a surrogate policy by minimizing the KL divergence between the policy and the policy constructed by the conservative Q. We evaluate our method on the Mujoco benchmark and demonstrate superior performance compared to previous state-of-the-art methods across various environments, particularly in the most complex Humanoid environment.
\end{abstract}

\section{Introduction}
Deep reinforcement learning(RL) has attracted much attention in recent years. It has achieved massive success in many fields, such as DQN \cite{mnih2015dqn} in simple RGB games, AlphaStar \cite{vinyals2019alphastar}, and OpenaiFive \cite{openai2019openaifive} in multi-player combat games, chatGPT \cite{chatgpt} in natural language processing. When applying deep RL in continuous action control, such as robotic control, there exist higher demands on the robustness of reinforcement learning policy \cite{haarnoja2017energy}. Algorithms based on the maximum entropy framework \cite{ziebart2010modeling} are more robust due to the diverse action selection, which augments the standard reward with the policy entropy, to some extent, encourages exploration in training and finally derives a robust policy. The intuitive reason for taking exploratory actions is that other actions with lower predicted rewards may be better.
Moreover, the method used to select actions directly affects the rate at which the RL algorithm will converge to an optimal policy. Ideally, the system should perform a non-greedy action if it lacks confidence in the current prediction. The RL method should perform a more greedy exploration once we gather more information about the prediction result.

Although various exploration methods, such as $\epsilon$-greedy, softmax, UCB-1~\cite{auer2002finite}, have been suggested for use in discrete action space, these kinds of explorations are not the same thing as the exploration in the continuous action space, due to the infinite actions. Since the actions in continuous space are uncountable, the exploration of actions is usually designed roughly, such as adding the Gaussian perturbation~\cite{Silver2014dpg,van2016ddqn,fujimoto2018addressing,haarnoja2018soft}. Intuitively, this kind of \textbf{aimless exploration} should not be the optimal exploration strategy. It will slow down learning the optimal policy due to its large randomness. We are not the first to consider this problem. The Optimistic Actor Critic~\cite{ciosek2019oac} method proposes exploring along the gradient direction of maximizing the Q function in continuous action space. When the estimated gradient direction is consistent with the real gradient direction of the Q function, it is equivalent to optimizing a step in advance. In this case, the OAC algorithm can achieve better results. But we found that the policy gradient cannot always be accurately estimated, resulting in a bad result. Another problem persists in the Q value-based policy gradient method since the DPG~\cite{Silver2014dpg} algorithm is proposed. That is, the policy learning is decoupled from the Q function learning. Since the Q function is related to action, the chain rule can be used to calculate the gradient of Q function with respect to action, thereby guiding the policy update. In this setting, at the same state, the policy can be much different from the action probability constructed by the Q function, \textbf{policy divergence} occurs, which heavily influences policy learning.
In short, these methods do not utilize the information learned by the Q function and optimize the action conservatively due to the overestimation of Q values without considering whether the overestimation is good or bad.

Assuming we can sample action with high confidence, it is more effective not to restrain the agent from choosing this action. From this intuition, based on the double Q estimation and soft policy learning, we design an exploration strategy that effectively explores the action space with the information provided by the Q function. In detail, we first propose the greedy Q softmax update, a precondition for our exploration strategy, and give the convergence analysis. Then, based on the above results, we propose a novel exploration strategy that combines the exploration with the update of the Q value based on the premise that the policy is consistent with the action probability distribution constructed by the Q function. We can update the Q value by sampling action from the distribution constructed by the greedy Q value instead of taking the argmax of the Q function, which is impracticable in continuous action space. Finally, to learn a policy that satisfies the mentioned premise, we make the policy learn from the Q function by minimizing the KL divergence between the policy and the distribution constructed by the conservative Q function. 

We evaluate our proposed method on Mujoco~\cite{todorov2012mujoco} benchmarks and verify that the proposed method outperforms the previous state-of-the-art in various environments, particularly the most complex Humanoid environment. We achieve about 8k scores in 3 million steps, a massive improvement over previous methods. Our method is related to soft Q learning, in which, assuming the shape of the Q function is multi-modal, we visualize the actual shape of the Q function in the Swimmer Environment(two action dimensions). In the training process, the visualized result shows that the Q function is multi-modal. We also discuss the two main issues raised in the paper in the experimental section. Finally, to further improve the practical usability of the method, we provide complete empirical and numerical results in the appendix for the three hyper-parameters.

\section{Preliminary}
We first introduce notation and the maximum entropy objective, then summarize the Soft Policy Learning method.
\paragraph{Notation}
In this paper, we consider deterministic policy reinforcement learning method for continuous action space.
Consider a discounted infinite-horizon Markov decision process (MDP), defined by the tuple $(\sspace, \aspace, \pdyn, \reward, \discount)$, where the state space $\sspace$ and the action space $\aspace$ are continuous, and the state transition probability $\pdyn:\ \sspace  \times \aspace \times \sspace \rightarrow [0,\, \infty)$ represents the probability density of the next state. Given the state $\st\in\sspace$ and action $\at\in\aspace$ at time-step $t$, we can get the probability density of $\stp\in\sspace$. The environment emits a bounded reward $\reward: \sspace \times \aspace \rightarrow [\rmin,\rmax ]$ on for specific state and action pair. 
$\discount$ is the discount factor, and its value is in the range $[0,1)$, which makes the infinite accumulated reward finite in mathematics.
\paragraph{Maximum entropy objective} Standard RL algorithm maximizes the expected sum of rewards $\sum_t \E{(\st,\at)\sim\rho_\policy}{\reward(\st,\at)}$. 
$\rho_\policy(\st,\at)$ denotes state-action marginals of the trajectory distribution induced by a policy $\policy(\at|\st)$.
Maximum entropy objective augment the expectation with the expected entropy of the policy over $\rho_\policy(\st)$:
\begin{align}
\label{eq:maxent_objective}
J(\policy)  =  \E{\policy}{\sum_{t=0}^{\infty} \reward(\st,\at) + \alpha\ent(\policy(\voidarg|\st))}.\notag
\end{align}
The temperature parameter $\alpha$ balance the relative importance of the entropy term and the reward, and this entropy term influence the exploration of the policy, which in result to a more stochastic optimal policy ideally.

\paragraph{Soft policy learning} Soft policy maximizes the maximize entropy objective and modifies the Q value function using the standard Q value function minus the current action's log probability, this Q value is called Soft Q value.
Considering the discount factor in practice algorithm, the standard Q value function is $ \E{(\st, \at) \sim \rho_\policy} { \sum_{t=0}^{\infty} \discount^t\reward(\st,\at)}$. The soft Q value is 
$\sum_{t=0}^{\infty} \E{(\st, \at) \sim \rho_\policy} {\discount^t \reward(\st,\at) +\alpha \discount^{t+1} \ent(\policy(\voidarg|\stp))}$.
For a fixed policy, the soft Q value can be computed iteratively, starting from any function $Q: \sspace\times \aspace \rightarrow \reals$ and repeatedly applying the modified Bellman backup operator $\mathcal{T}^\policy$ given by
\[
\mathcal{T}^\policy Q(\st, \at) \triangleq  \reward(\st, \at) + \discount \E{(\stp,\atp) \sim \rho_\policy}{\Q(\stp, \atp) - \alpha\log\policy(\atp|\stp)},
\]
then improve the policy by minimizing following formula
\[
\policy' = \arg\underset{\policy\in \Pi}{\min}\kl{\policy(\voidarg|\st)}{\frac{\exp\left(Q(\st, \voidarg)\right)}{Z(\st)}},
\]
where $Z(\st)=\sum_{\at}Q(\st,\at)$ normalizes the distribution.

\section{Issues in previous work}%
As mentioned in the introduction, modern continuous RL methods explore inefficiently.  We now describe two phenomena to explain why it is inefficient.
\begin{figure}[htbp]
    \centering
    \subfigure[Aimless exploration]{
    \label{fig:aimless_exploration}
        \includegraphics[width=0.45\textwidth, trim={0 0 4mm 0mm}, clip]{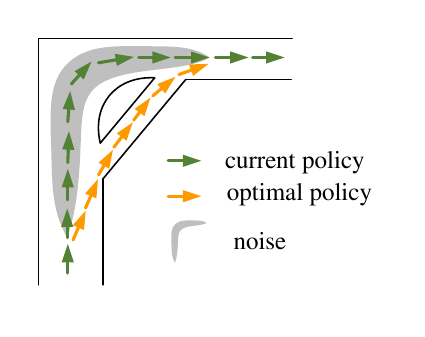}
    }
    \subfigure[Policy divergence]{
    \label{fig:policy_divergence}
    \includegraphics[width=0.45\textwidth, trim={0 0 4mm 0mm}, clip]{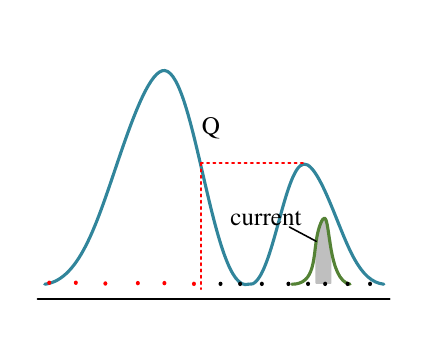}
    }

    \caption{Two phenomena of Inefficiency Exploration. Left: Exploration is usually based on the current policy and implemented by adding noise or perturbation. Right: The Q function and policy $\pi$ are learned separately, and policy learning often lags, causing policy divergence. Red points should be sampled instead of black points to avoid falling into a sub-optimal policy.} 
    \label{fig:two_issue}
\end{figure}

\paragraph{Aimless exploration}

Aimless exploration refers to a form of exploration in which an agent takes random actions without a clear goal. This type of exploration can be inefficient and time-consuming, as the agent may spend significant amounts of time exploring unimportant or irrelevant areas of the environment. As shown in Figure~\ref{fig:aimless_exploration}, when the current policy is poorly initialized and far away from the optimal policy, exploring the optimal policy without a specific objective in mind can be challenging. This aimless exploration is inefficient and leads to poor performance, as the agent may fail to discover important states or actions necessary for achieving its objectives. Exploration that relies solely on the current policy is akin to performing a depth-first search at a state, which is limited by the quality of the policy initialization and the difficulty of improving the policy. Therefore, it is beneficial to construct a policy that can guide exploration. Guided exploration is similar to performing a breadth-first policy search at a state, which can help address the issues associated with the current policy-based approach.

\paragraph{Policy divergence}
The phenomenon of the policy $\pi$ away from its target policy (constructed by the Q function) is referred to as policy divergence. In the deterministic policy gradient method, since the policy learning is separate, the Q function is no longer used to make decisions, so the policy always lags. When we optimize the policy, we may face similar issues to the out-of-distribution (OOD) problem encountered in offline reinforcement learning. The typical training process involves sampling actions within the policy's gray area, as shown in Figure~\ref{fig:policy_divergence}. Then evaluate the corresponding Q values and update the policy parameter that maximizes the Q function. Even if we sample far from the current policy(such as explicit sampling from OOD points), it may not be effective, as shown by the black sampling point in the figure. To learn the optimal policy, we must be able to sample the red points as more as possible. Sampling from OOD points is core to solving this issue. Policy divergence can result in the policy failing to improve and, in some cases, making worse estimates due to the compromised optimal action selection.


\section{Improving exploration for soft policy learning}
In this section, we will first introduce a novel Q value update method. Next, we construct an effective exploration strategy and combine the value update and action exploration based on a premise. Then, we show how to satisfy the premise and learn an effective policy. 
\subsection{Greedy Q softmax update}\label{sec:basis}
In this section, we propose the greedy Q operator(GDQ) for value function updates. This method is based on the double Q estimation, it uses two separate Q functions to estimate the value of state-action pairs. "greedy" means it always selects the maximum Q values of these two Q functions. We first define the greedy Q function,
\[
Q^{max}(s, a) = \max \{Q^1 (s, a),Q^2 (s, a)\},Q^{min}(s, a) = \min \{Q^1 (s, a),Q^2 (s, a)\},
\]
then GDQ operator is defined as follows: for $\forall s \in \sspace$,
\[
gdq_{\beta_t} (Q(s,\voidarg))=\frac{\sum_{a \in \aspace} e^{\beta_t Q^{max} (s,a)} Q^{min}(s,a)}{\sum_{a \in \aspace} e^{\beta_t Q^{max}(s,a)}}
\]
where $\beta_t$ is a dynamically increased hyper-parameter during the training iteration. We now give theoretical analysis of the proposed GDB operator and show that it has good convergence guarantee.

A modified Bellman backup operator $\mathcal{T}^\policy$ given by
\begin{align}
\notag
\mathcal{T}^\policy Q(\st, \at) \triangleq  \reward(\st, \at) + \discount \E{\stp \sim \pdyn}{V(\stp)},
\end{align}
where
\begin{align}
V(\st) = gdq_{\beta_t} (Q(s,\voidarg))
\notag
\end{align}
\begin{theorem}[Convergence of value iteration with the GDQ operator]
\label{the:Convergence_of_GDQ}
For any dynamic greedy Q operator $gdq_{\beta_t}$, if $\beta_t$ approaches $\infty$ after $t$ iterations, the value function  $Q_t$ converges to the optimal value function $Q^*$.
\end{theorem}

The proof is deferred to Theorem~\ref{app:the_Convergence_of_GDQ}. We extend the use of the DBS operator~\cite{bloz2020pan} based on greedy Q, which is less affected by overestimation. The motivation is that in continuous action space, the maximization of the value function suffers from overestimation, and taking the argmax of the value function is impractical. It is a better way that sample finite action to estimate the target Q value and improve policy iteratively. 

\subsection{Exploration with greedy Q }\label{sec:exp}
Based on the above content, we propose a novel exploration strategy. 
We first define the exploration policy $\pi_{E}$,
\[
\pi_{E}(\voidarg|\st)=\frac{e^{\beta_t Q^{max} (\st,\voidarg)}}{\sum_{a \in \aspace} e^{\beta_t Q^{max}(\st,a)}}.
\label{eq-exploration}
\]
According to the result of the Theorem~\ref{the:Convergence_of_GDQ}, we can use the following formula to update the target Q value:
\begin{equation}
\label{eq:gdq_Q}
\reward(\st, \at) + \discount \E{\stp \sim \pdyn, \atp \sim\pi_{E}(\voidarg|\stp)}{\Q(\stp, \atp)}.
\end{equation}
However, computing the target next state values is computationally expensive. It needs to sample over all possible states and actions and then compute the corresponding Q-values. Refer to the SARSA method, we can sample two consecutive (s,a) pairs to estimate the expectation of the Q value:
\begin{align}
\label{eq:sarsa_gdq_q}
  \E{\st \sim \pdyn ,\at \sim \pi_{E}(\voidarg|\st) ,\stp\sim\pdyn,\atp \sim \pi_{E}(\voidarg|\stp) }{\reward(\st, \at) +\discount Q(\stp, \atp)}.
\end{align}
The main difference is that now we can estimate the expectation of Q value with finite sampling. The target Q value(\eqref{eq:gdq_Q}) requires evaluating the next Q value in the entire state and action space. But the consecutive pairs require the computation in an on-policy form. For continuous action RL task, we learn the policy separated from the Q function. If we can sample the action from the policy $\pi$ as follows:
\begin{align}
\label{eq:gdq_exp_f_target}
  \E{(\st,\at,\stp) \sim (\pdyn,\pi_{E}(\voidarg|\stp),\pdyn)}{ \reward(\st, \at) + \discount \E{\atp \sim \pi(\voidarg|\stp)}{\Q(\stp, \atp)}},
\end{align}
then we can use this equation to update the Q function in an off-policy form. It can see that we have obtained a new exploration strategy.

\begin{figure}[htbp]
    \centering
    \subfigure[Double Q]{
        \includegraphics[width=0.42\textwidth, trim={0 2mm 0mm 2mm}, clip]{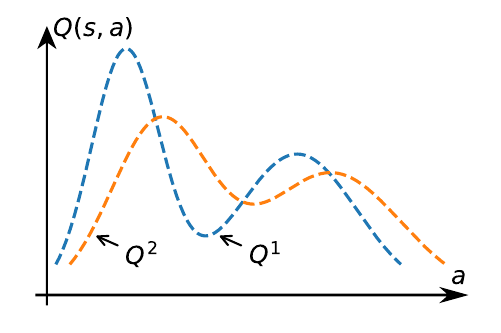}
    }
    \hfill
    \subfigure[Greedy Q]{
    \includegraphics[width=0.42\textwidth, trim={0 2mm 0mm 2mm}, clip]{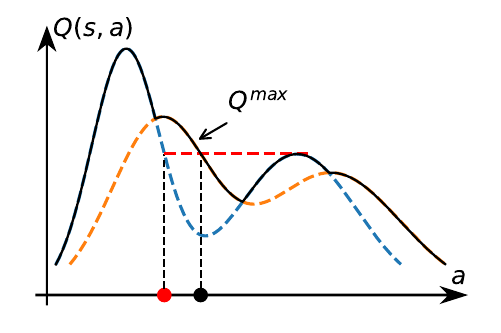}
    }
    \caption{Visualization of the Q function. The state $s$ is fixed. Left: The two Q functions are in an energy-based form, which is the optimal solution for the maximum-entropy objective. Right: Greedy Q function take the max value of these two Q functions over the action space. The range of the red point becomes large if we sample according to value of the Greedy Q instead of the $Q^1$.} 
    \label{fig:greedy_double_q_exp}
\end{figure}

As shown in Figure~\ref{fig:greedy_double_q_exp}, our proposed greedy Q exploration strategy has the following advantages:
1) It is better for exploration than any single Q function. As the black point and red point shown in the figure, the number of action, which is better than the sub-optimal action, increases, and relative range is larger. 
2) The max operator in our method is also one kind of overestimation. Overestimation is awful for Q-value update, but relative good is a better property when using it for exploration. 
3) While our method is named 'Greedy', it actually encourages exploration by reducing the probability of selecting the action with the highest value. This is achieved by overestimating the values of all available actions. 

We also need discuss the prerequisite for the replacement from Equation~\eqref{eq:sarsa_gdq_q} to Equation~\eqref{eq:gdq_exp_f_target}. It require the action sample from the distribution $\pi(\voidarg|\stp)$  as consistent as possible with the action sampled from the distribution $\pi_{E}(\voidarg|\stp)$. That is, to ensure that these two policy are as consistent as possible. Next, we discuss how to learn this policy $\pi$.

\subsection{Policy learning}\label{sec:learning}

The target action distribution for optimization is defined as follows:
\[
\pi_{O}(\voidarg|\st)=\frac{e^{ Q^{min} (\st,\voidarg)}}{\sum_{a \in \aspace} e^{ Q^{min}(\st,a)}},
\]
where the conservative Q is defined as $Q^{min}(\st, \at) = \min \{Q^1 (\st, \at),Q^2 (\st, \at)\}$, then we can make the policy directly learn from the target policy like the soft policy learning as follows:
\begin{align}
    \policy'
    &= \arg\underset{\policy}{\min}\kl{\policy(\voidarg|\st)}{\pi_{O}(\voidarg|\st)},\notag
\end{align}
Now consider the neural network parameterized $Q_\theta$ function and policy $\pi_\phi$, thus,
\begin{equation}
    Q^{max}(\st, \at) = \max \{Q_{\theta_1} (\st, \at),Q_{\theta_2} (\st, \at)\},Q^{min}(\st, \at) = \min \{Q_{\theta_1} (\st, \at),Q_{\theta_2} (\st, \at)\}
\end{equation}
Then we can learn the policy by minimizing the expected KL-divergence policy objective,
\begin{align}
J_\pi(\phi) &= \E{\st\sim\mathcal{D}}{\kl{\pi_\phi(\voidarg|\st)}{\pi_{O}(\voidarg|\st)}} \notag \\ 
&=\E{\st\sim\mathcal{D}}{\kl{\pi_\phi(\voidarg|\st)}{ \exp (Q^{min}(\st,\voidarg) - \log Z_\theta(\st)) }} \notag \\ 
&=\E{\st\sim\mathcal{D}}{\E{\at \sim \pi_\phi(\voidarg|\st)} { \log \pi_\phi(\at|\st) - Q^{min}(\st,\at) + \log Z_\theta(\st) }},\label{eq:policy_repara}
\end{align}
where $Z_\theta(\st)= \sum_{a \in \aspace} \exp{ Q_{\theta}^{min}(\st,a)}$, $\mathcal{D}$ is a replay buffer and Equation~\eqref{eq:policy_repara} requires sampling action from the policy distribution. The re-parameterized trick is used as
\begin{equation}
  \at = f_\phi(\epsilon_t; \st), \epsilon_t \sim \gauss(\mu,\sigma^2). \label{eq:gauss}
\end{equation}
The gradient of $J_\pi(\phi)$ with respect to $\phi$ as follows:
\begin{align}
\nabla_\phi J_\pi(\phi) &= \nabla_\phi \E{\st\sim\mathcal{D}, \epsilon_t \sim \gauss } { \log \pi_\phi(\at|\st) - Q^{min}(\st,\at)} \notag\\
&= \E{\st\sim\mathcal{D}, \epsilon_t \sim \gauss } {\nabla_\phi \log \pi_\phi(\at|\st) - \nabla_\phi Q^{min}(\st,\at)|_{\at=f_\phi(\epsilon_t; \st)} },\label{eq:gradient_objective}
\end{align}
where the computation of $Z_\theta(\st)$ is independent of the $\phi$ and because we use neural network to parameterized the policy and Q function, for the two term in Equation~\eqref{eq:gradient_objective}, we can use deep leaning framework to finish the forward computation, the auto gradient mechanism will finish the back-propagation by the framework itself. And we can derive the unbiased estimation of Equation~\eqref{eq:gradient_objective} with the following equation
\begin{equation}
    \hat{\nabla}_\phi J_\pi(\phi) = \nabla_\phi \log \pi_\phi(\at|\st) - \nabla_\phi Q^{min}(\st,\at)|_{\at=f_\phi(\epsilon_t; \st)}.
\end{equation}

Here, refer to Equation~\eqref{eq:gdq_exp_f_target}, we also give the Q learning objective:
\begin{align}
  J_\Q(\theta) = \E{(\st,\at) \sim \mathcal{D} }{ 
  \frac{1}{2}( Q_\theta(\st,\at) - \hat{Q}(\st,\at) )^2
  },\label{eq:q_obje}
\end{align}
where $\hat{Q}(\st,\at) = \reward(\st, \at) + \discount \E{\epsilon_{t+1} \sim \gauss}{\Q^{min}(\stp, \atp)- \log \pi(\atp|\stp)}$, and $\atp=f_\phi(\epsilon_{t+1}; \stp)$, the $- \log \pi(\atp|\stp)$ term is due to the computation is based on the maximum entropy framework and the Q function is also in an energy-based form. 
The transition from replay buffer $\mathcal{D}$ is generated from the interaction of the policy $\pi_{E}$ and the environment. Then the gradient of the Q learning objective(Equation~\eqref{eq:q_obje}) can be estimated with an unbiased estimator
\begin{align}
\hat{\nabla}_\theta J_Q(\theta) =  \nabla_\theta \Q_\theta(\at, \st) \left(\Q_\theta(\st, \at) - \reward(\st, \at) - \discount \Q^{min}(\stp, \atp)+ \discount\log \pi(\atp|\stp) \right). \notag
\end{align}

\subsection{The greedy exploration algorithm}\label{sec:gac_algo}

Greedy Actor-Critic (see Algorithm~\ref{alg-gac} in the appendix) dynamic increases $\beta_t$ to guarantee the convergence  of Q value(line~\ref{alg-line-beta_t} as described in section~\ref{sec:basis}), then samples action from the exploration policy(line \ref{alg-line-explore} to interact with the environment, as described in section~\ref{sec:exp}) and stores the transition in a memory buffer, finally, GAC samples transitions from the memory buffer to update the Q function (line \ref{alg-line-q}) and the actor (line \ref{alg-line-actor}) as described in section~\ref{sec:learning}. In detail, the policy network outputs the $\mu$ and the $\sigma$ of Equation~\eqref{eq:gauss}. We uniform sample $s_n$ actions from the range $[\mu-s_r*\sigma, \mu+s_r*\sigma]$, then evaluate sampled actions to construct the $\pi_{E}$ distribution, $s_r$, $s_n$ and $\beta_t$ is the three main hyper-parameter in our algorithm. We will discuss the time cost in section \ref{sec-disc}.
\subsection{Related work}
In this section, we discuss topics related to continuous action RL algorithms.

\paragraph{Exploration}
The classical exploration methods in value-based reinforcement learning are $\epsilon$-greedy, softmax, and UCB-1~\cite{auer2002finite}. In policy gradient methods, exploration can be achieved by utilizing information from the policy, such as entropy. The deterministic policy gradient~\cite{Silver2014dpg} method separated policy learning from Q-function learning. Since then, deterministic-policy-based methods explored by random sampling action around the current policy until the Optimistic Actor-Critic~\cite{ciosek2019oac} method appears. It considers combining two Q-functions for exploration by sampling around the predicted better action based on the policy gradient. However, we found that it is also limited to accurately estimating policy gradients. Thus, we propose a more direct method: using the Q-function to evaluate actions for exploration.

We also note that some heuristic exploration methods for continuous RL tasks emerge, such as the Coherent Exploration~\cite{ex-Coherent-zhang21t} algorithm, directly modify the policy network's last layer parameters to enhance the policy's exploratory nature. The DOIE~\cite{ex-optinitial} algorithm explores using a modified Q-function, which involves assigning an optimistic value to transitions that lie considerably beyond the agent's prior experience. The RRS~\cite{rrs_sun_2022} algorithm directly alters Q values, which can be viewed as adjusting the initialization parameters of the Q network. This modification contributes to increased exploration diversity.

\paragraph{Overestimation}
The concept of overestimation was first introduced in the paper by Thrun and Schwartz~\cite{thrun1993issues}, discussing the positive approximation error in the function approximation for RL. Then the MCQ-L~\cite{rummery1994sarsa} method(famous with the name ``SARSA''\cite{sutton2018reinforcement}) mentioned that the argmax operator is impractical in training. They estimate the Q value with the consequent two-state-action pairs(in an online form). The Double Q learning~\cite{hasselt2010dql} updates the Q value with two estimators to avoid overestimation. Then the Double DQN ~\cite{van2016ddqn} is proposed, which parameterizes the Q function with a neural network. Inspired by the Double DQN,  TD3~\cite{fujimoto2018addressing} algorithm is proposed and uses double Q values and delayed Q update to alleviate the overestimation. Our approach utilizes overestimation to guide exploration, as the overestimated actions are generally considered relatively good.

\paragraph{Policy learning}
Stochastic policy gradient methods, such as A3C~\cite{mnih2016asynchronous}, TRPO~\cite{schulman2015trust}, and PPO~\cite{schulman2017proximal}, can be used for policy learning in continuous action spaces. However, optimizing stochastic policy gradients in continuous action spaces is challenging, value-based deterministic policy gradient methods get better result. In the DPG~\citep{Silver2014dpg} algorithm,  the policy parameter is optimized toward maximizing the Q function. In the DDPG~\cite{lillicrap2016continuous} algorithm, the Q function is parameterized by a neuron network. Then the SQL~\citep{haarnoja2017energy} algorithm is proposed, which assumes the Q function has an energy-based form. The  TD3~\cite{fujimoto2018addressing} algorithm first proposes updating the policy parameter conservatively with the minimum value of the two Q functions. And then, the SAC\cite{haarnoja2018soft} algorithm uses a stochastic actor for policy exploration and optimization. Essentially, these methods do not improve how the policy is optimized, nor do we. We accelerate the learning of the policy through effective exploration.

\section{Experiment}\label{sec-disc}
We conducted experiments using the Mujoco physics engine~\cite{todorov2012mujoco}, which is currently free to use and is maintained by Deepmind. In the following, we present our experimental findings and analyses. More detailed results are in the appendix~\ref{sec:more_results}.
\begin{figure}[ht]
    \centering
    \includegraphics[width=0.98\linewidth]{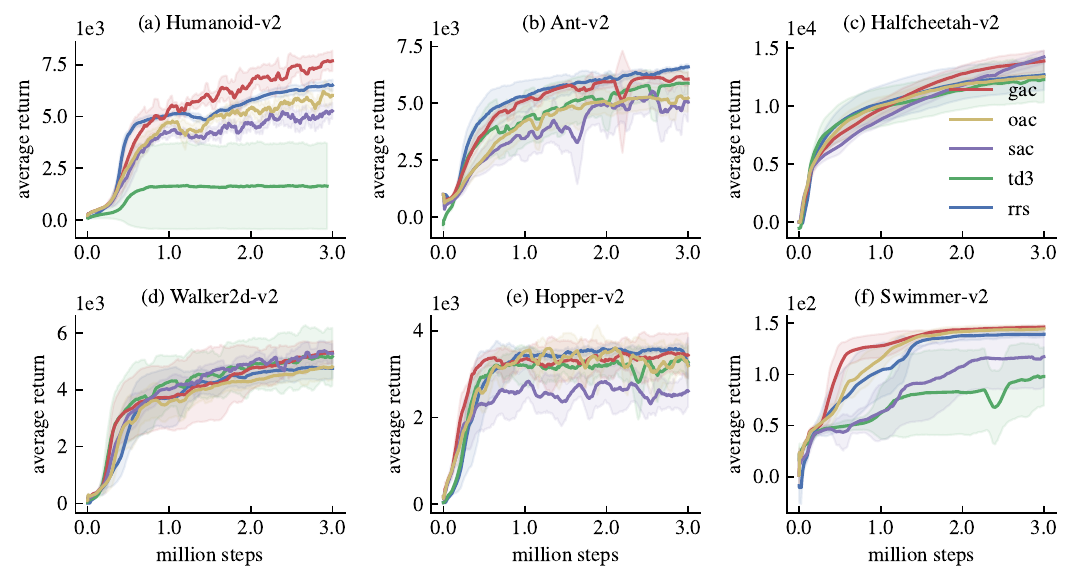}
    \caption{Results of GAC and four baseline algorithms in the six continuous environments}
    \label{fig:overall_result}
\end{figure}

\paragraph{General results on MuJoCo benchmark} We compare GAC to OAC[2019]~\cite{ciosek2019oac}, SAC[2018]~\cite{haarnoja2018soft}, TD3[2018]~\cite{fujimoto2018addressing} and RRS[2022]~\cite{rrs_sun_2022} , four recent model-free RL methods that achieve state-of-the art performance. All methods run with six random seeds. The policy network and the Q network are the same for all methods. GAC uses three hyper-parameter related to exploration, which has been introduced in section~\ref{sec:gac_algo}. We provide the value of all hyper-parameter in the appendix~\ref{app:hyper_parameter}. The results are organized based on the complexity of the environment, ranging from complex to simple, as illustrated by Figure 3(a) through Figure 3(f). The Humanoid environment is the most complex, and the Swimmer environment is the simplest. The state dim and action dimension are summarized in the appendix~\ref{app:environment_properties}. As shown in Figure~\ref{fig:overall_result}, our method achieves promising results on this benchmark. On Humanoid-v2, GAC achieves state-of-the-art performance and is sample efficient than previous algorithms. On Ant-v2, GAC works slightly worse than the RRS algorithm in the final performance. On Halfcheetah-v2, our method get better sample efficient. On Walker2d-v2 and Hopper-v2, our method get similar results with others. On Walker2d-v2, our method work better in the early learning stage.

\paragraph{Visualization of the Q function} 
\begin{figure}[htbp]
    \centering

    \subfigure[Swimmer]{
        \includegraphics[width=0.28\textwidth, trim={0 -4mm 0mm 6.3mm}, clip]{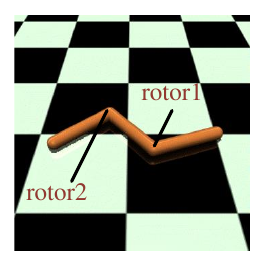}
    }
        \subfigure[3D visualization]{
        \includegraphics[width=0.34\textwidth,  trim={0 -0.1 0 0}, clip]{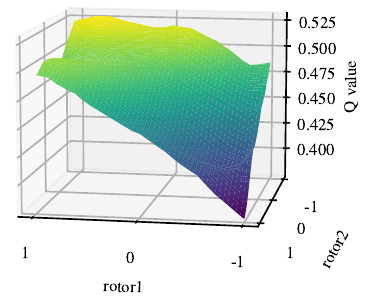}
    }
        \subfigure[2D plane]{
        \includegraphics[width=0.32\textwidth, trim={0 0.1mm 0mm 0.1mm}, clip]{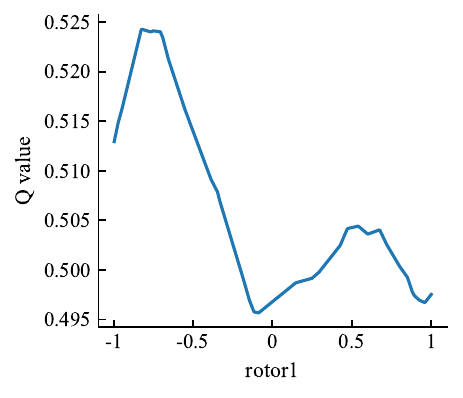}
    }

    \caption{Visualization of the Q function. (a): The swimmer has two rotors, and its moving is controlled by adjusting the torque applied to the two rotors. (b): The Q values of the two-dimension actions are plotted in 3D Space. (c): We plot a particular case for rotor2=-1 to show that the Q function has an energy-based form in early-stage training.} 
    \label{fig:visulization_Q}
\end{figure}
Our approach is based on the maximum entropy framework and assumes that the Q function is in an energy-based form. We design experiments to validate this assumption. The action space in the Simmer environment is two-dimensional, making it an ideal validation environment. We select an intermediate state of the Q network during the training process, sample 400*400 points across the entire action space, and calculate the corresponding Q values. The results we obtained are shown in Figure~\ref{fig:visulization_Q}. We plot the 3d surface of the Q function, and a 2d plane for rotor2=-1.

\begin{figure}[htbp]
    \centering
    \subfigure[Hyper-parameter $\beta$]{
    \label{fig:beta}
        \includegraphics[width=0.3\textwidth, trim={0 0 0mm 0mm}, clip]{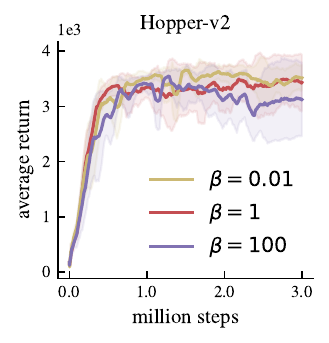}
    }
        \subfigure[Sample range]{
        \label{fig:range}
        \includegraphics[width=0.3\textwidth, trim={0 0 0mm 0mm}, clip]{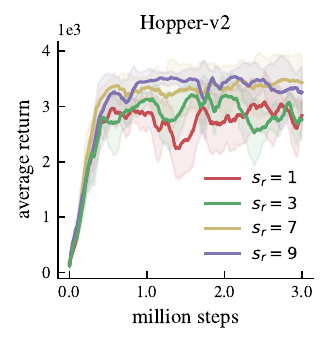}
    }
        \subfigure[Sample size]{
        \label{fig:number}
        \includegraphics[width=0.3\textwidth, trim={0 0 0mm 0mm}, clip]{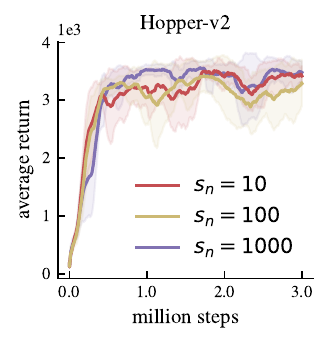}
    }

    \caption{Three hyper-parameters related to the exploration policy. (a): In practice, the value of $\beta_t$ is obtained by multiplying $\beta$ with the epoch number of timestep $t$. (b): The parameter $s_r$ determines the sample range, where a large value indicates that sampled actions could deviate further from the distribution of the current policy. (c): We uniformly sample $s_n$ actions within the sample range to construct our exploration policy.} 
\end{figure}

\paragraph{Learning the Q value}
The parameter $\beta_t$ influences the learning of the Q function. In practice, we dynamically increase the $\beta_t$ by setting it as the multiplication of  $\beta$ and the epoch number of timestep $t$. In the early stage of training, the Q function cannot provide little information, so a small $\beta$ can be used to encourage the exploration. The different $\beta$ results are shown in Figure~\ref{fig:beta}. As we can see, a smaller $\beta$ can produce a slightly better result. Our method is not so sensitive to this parameter in our setting. Nevertheless, a smaller $\beta$ indeed takes a better result.
\paragraph{Better exploration}
As shown in Figure~\ref{fig:range}, if we use a smaller value of $s_r$, that is, we sample action around the current policy, we can see the final result compared to the two larger values is terrible, which shows that \textbf{aimless exploration} does lead to poor results, it is difficult to get good results if $s_r$ is too small. Policy learning depends on how well the initial policy is. When $s_r$=7, we get better results indicating that the \textbf{policy divergence} problem can be solved by explicit sampling point out-of-distribution. Thus, we can do more  OOD sampling with the Q function to explore action space. Additionally, according to the results of $s_r$=7 and $s_r$=9, we know that $s_r$ should not be as big as possible. Bigger $s_r$ does not mean better results because the policy gradient comprises the $\nabla_\phi Q$ and $\nabla_\phi\pi$. Suppose we optimize the policy with many low-probability actions, which may prefer by the Q function. In that case, the policy gradient may be too small to promote the parameter update.

\paragraph{Exploration policy}
When constructing an exploration policy, a key question is how many actions need to be evaluated to obtain a usable exploration policy. If constructing the policy needs many actions to be evaluated, our method becomes computationally burdensome and resource-intensive. As shown in Figure~\ref{fig:number}, when $s_n$=10, the final result is the best; when $s_n$=1000, the best result is achieved at about one million steps, but the final result does not gain an advantage. According to the results in the figure, it can be seen that GAC can be effective by evaluating only a limited number of actions, and the improvement of the final performance does not lie in the complete evaluation of actions in the action space. Instead, the exploration strategy is the main factor for performance improvement.

\paragraph{Time cost}Our experimental setup consists of a computer with Ubuntu 18 operating system, equipped with a 9900K CPU and an RTX 2060 GPU. Without using the exploration strategy proposed in this paper, it takes 6.619 seconds to complete one epoch on average. In most of our experiments, we evaluate 32 actions. With this setting, running one epoch takes an average of 7.29 seconds. If we evaluate 64 actions, running one epoch takes an average of 7.09 seconds. The GPU may be more efficient when computing data with a batch size 64. From this, the additional time added due to exploration is insignificant. One epoch requires 1000 interactions with the environment. When averaged per exploration step, the time consumed is almost negligible.

\section{Conclusion}
In this paper, we derive a practical exploration strategy for deterministic policy reinforcement learning in continuous action space based on the greedy softmax Q update theorem, which leverages the advantages of the double Q estimate and utilizes information provided by the Q function. Compared with previous methods, using greedy Q value for exploration achieves faster policy learning. We do experiments on the Mujoco benchmark. Our method significantly improves over previous methods in complex environments, and empirical results show that we improve exploration for the deterministic policy gradient method.

\bibliographystyle{plain}
\bibliography{eefmef}

\newpage
\appendix
\Large\textbf{Appendix}
\section{Proofs}
\normalsize 
\subsection{Theorem~\ref{the:Convergence_of_GDQ}}
\label{app:the_Convergence_of_GDQ}

This proof (of Theorem 1) uses techniques from the proof of Theorem 1 in the paper~\cite{bloz2020pan}, adapting them to the setting considered in this paper. An informal overview is as follows. The maximum of the two Q values is a particular case in the computation of the operator. Here, we mainly show this process also can guarantee convergence.
We start with a proposition from the paper~\cite{bloz2020pan} that shows the relation between the soft-max and the log-sum-exp function.  

\textbf{Proposition 1}
\emph{
	\begin{equation}
	\text{lse}_{\beta}({\bf{X}}) - \text{sm}_{\beta} ({\bf{X}}) =\frac{1}{\beta}H({\bf{X}})= \frac{1}{\beta} \sum_{i=1}^n -p_i \log(p_i) \leq \frac{\log(n)}{\beta},
	\end{equation}
	where $p_i = \frac{e^{\beta x_i}}{\sum_{j=1}^n e^{\beta x_j}}$ denotes the weights of the softmax distribution, $\text{lse}_{\beta}({\bf{X}})$ denotes the log-sum-exp function $\text{lse}_{\beta}({\bf{X}}) = \frac{1}{\beta} \log(\sum_{i=1}^n e^{\beta x_i})$, and $\text{sm}_{\beta}({\bf{X}})$ denotes the softmax function $\text{sm}_{\beta}({\bf{X}})=  \frac{\sum_{i=1}^n e^{\beta x_i}x_i}{\sum_{j=1}^n e^{\beta x_j}}$. $H({\bf{X}})$ is the entropy of the distribution. It is easy to check that the maximum entropy is achieved when $p_i = \frac{1}{n}$, where the entropy equals to $\log(n)$.
}

\begin{proof}
	\begin{align*}
	&\frac{1}{\beta} \sum_{i=1}^n -p_i \log(p_i) \\
	= &\frac{1}{\beta} \sum_{i=1}^n \left( -\frac{e^{\beta x_i}}{\sum_{j=1}^n e^{\beta x_j}} \log \left(\frac{e^{\beta x_i}}{\sum_{j=1}^n e^{\beta x_j}} \right) \right) \\
	= &\frac{1}{\beta} \sum_{i=1}^n \left( -\frac{e^{\beta x_i}}{\sum_{j=1}^n e^{\beta x_j}} \left( \beta x_i - \log \left( \sum_{j=1}^n e^{\beta x_j} \right) \right) \right) \\
	= &-\sum_{i=1}^n \frac{e^{\beta x_i} x_i}{\sum_{j=1}^n e^{\beta x_j}} + \frac{1}{\beta} \log \left( \sum_{j=1}^n e^{\beta x_j} \right) \frac{\sum_{i=1}^n e^{\beta x_i}}{\sum_{j=1}^n e^{\beta x_j}} \\
	= &-\text{sm}_{\beta} ({\bf{X}}) + \text{lse}_{\beta}({\bf{X}})
	\end{align*}

\end{proof}
\textbf{Theorem 1 (Convergence of value iteration with the GDQ operator)} \emph{
For any dynamic greedy Q operator $gdq_{\beta_t}$, if $\beta_t$ approaches $\infty$, the value function after $t$ iterations $v_t$ converges to the optimal value function $V*$.
}
\begin{proof}
    \begin{align}
	& ||(\mathcal{T}_{\beta_t} V_i) - (\mathcal{T}_{\beta_t} V_j)||_{\infty} \\
	=& \max_s | {\rm{gdq}}_{\beta_t} (Q_i^{max}(s,\cdot)) - {\rm{gdq}}_{\beta_t} (Q_j(s,\cdot)) | \\
 	\leq & \max_s | {\rm{lse}}_{\beta_t}(Q_i^(s, \cdot)) - \frac{1}{\beta}H(Q_i(s, \cdot)) \notag \\
  &-{\rm{lse}}_{\beta_t} Q_j(s, \cdot))+\frac{1}{\beta}H(Q_j(s, \cdot)) |  \\
	\leq & \underbrace{\max_s | {\rm{lse}}_{\beta_t}(Q_i(s, \cdot)) -{\rm{lse}}_{\beta_t} ((Q_j(s, \cdot)) |}_{(A)} + \underbrace{\frac{\log(|A|)}{\beta_t}}_{(B)} , 
	\label{th1_2}
	\end{align}
For the term $(A)$, the log-sum-exp operator has been proved as a non-expanding operator in the paper \cite{fox2015taming}. In mathematics, the log-sum-exp function is almost equal to the max function. That is why this makes sense. Here, we prove that it's also a non-expanding operator in our greedy Q setting.
Define a norm on Q value as $\|\Q_i-\Q_j\| \triangleq \max_{\state,\action}|\Q_i(\state,\action) - \Q_j(\state,\action)|$. Suppose $\epsilon  = \|\Q_i-\Q_j\|$. Please note that in our setting,$\Q_i(\state,\action)=\min{\Q_i^1(\state,\action),\Q_i^2(\state,\action)}$, $\Q_j(\state,\action)$ is similar defined. Then 
\begin{align}
    \log \int \exp (\Q_i(\state',\action')) \ d\action' &\leq \log \int \exp( \Q_j(\state',\action')  + \epsilon )\ d\action'\notag\\
    &= \log \left(\exp(\epsilon )\int \exp \Q_j(\state',\action') \ d\action'\right) \notag\\
    &= \epsilon  + \log \int \exp \Q_j(\action',\action') \ d\action'.
\end{align}
Similarly, $\log \int \exp \Q_i(\state',\action') \ d\action' \geq -\epsilon + \log \int \exp \Q_j(\state',\action') \ d\action'$. Therefore $\|\mathcal{T}\Q_i - \mathcal{T}\Q_j\| \leq \discount \epsilon = \discount \|\Q_i - \Q_j\|$. 

Consider for $Q^1$ and $Q^2$, we expand the composed Q function, we derive
\begin{equation*}
\|\mathcal{T}\Q_i - \mathcal{T}\Q_j\| \leq \discount  \|\Q_i - \Q_j^{1}\| ,\|\mathcal{T}\Q_i - \mathcal{T}\Q_j\| \leq \discount  \|\Q_i - \Q_j^{2}\|
\end{equation*}
when $\Q_i=\Q_i^{1}$, it means $\Q_i^{1}$ is the min value, for the term $(A)$,we have
\begin{align*}
    &\max_s | {\rm{lse}}_{\beta_t}(Q_i(s, \cdot)) -{\rm{lse}}_{\beta_t} ((Q_j(s, \cdot)) |\\
    &\leq \max_{s,a} \frac{1}{\beta_t} |\beta_t Q_i^1 - \beta_t Q_j^1|\\
    &\leq \discount \max_{s,a} \sum_{s'}p(s'|s,a) |V_i(s')- V_j(s')|\\
    &\leq \discount ||V_i -V_j||_{\infty}
\end{align*}
So $\mathcal{T}$ is a contraction. We can get the same result when $\Q_i=\Q_i^{2}$. And due to the min operator, $Q^1$ and $Q^2$ are updated iteratively. Consequently, the two Q functions converge to the optimal value, satisfying the modified Bellman equation. Thus, the optimal policy is unique.
	
For the term $(B)$, the details can be found in \cite{bloz2020pan}. A direct understanding is that as $\beta$ increases, this term will eventually become zero.
\end{proof}
\section{Algorithm}
Algorithm~\ref{alg-gac} is the practice algorithm of our method.
\begin{algorithm}[ht]
\caption{Greedy Actor-Critic (GAC).}
\label{alg-gac}
\begin{algorithmic}[1]
\Require $\theta_1$, $\theta_2$, $\phi$ \Comment{Initial parameters $\theta_1,\theta_2$ of the Q function and $\phi$ of the target policy $\pi_T$.}
\State $\breve \theta_1 \leftarrow \theta_1$, $\breve \theta_2 \leftarrow \theta_2, \mathcal{D}\leftarrow\emptyset$ \Comment{Initialize target network weights and replay buffer}
\For{each iteration}
        \State increase $\beta_t$ according to the iteration number \label{alg-line-beta_t}
	\For{each environment step}

	    \State \label{alg-line-explore} $\at \sim \pi_E(\at|\st,\mathbf{\beta}_t)$ \Comment{Sample action from exploration policy as in \eqref{eq-exploration}.}
	    \State $\stp \sim p(\stp| \st, \at)$ \Comment{Sample state from the environment}
	    \State $\mathcal{D} \leftarrow \mathcal{D} \cup \left\{(\st, \at, R(\st, \at), \stp)\right\}$ \Comment{Store the transition in the replay buffer}
	\EndFor
	\For{each training step}
        \State sample batch transition $(\st, \at, R(\st, \at), \stp)$ from the buffer
        \State \label{alg-line-v-compute} $\text{compute target } \hat{Q}^i(\st,\at)$ for $i \in {1,2}$
	     \State \label{alg-line-q} $ \text {update} \; \theta_i  \; \text{with}$ $  \hat \nabla_{\theta_i} J_{Q}(\theta_i)$  for $i \in {1,2}$ \Comment{Q parameter update }
	     \State \label{alg-line-actor} $ \text {update} \;\phi  \; \text{with} \; \hat \nabla_\phi J_\pi (\phi) $ \Comment{Policy parameter update}
	\State $\breve{\theta}_1\leftarrow \tau \theta_1 + (1-\tau)\breve{\theta}_1, \breve{\theta}_2\leftarrow \tau \theta_2 + (1-\tau)\breve{\theta}_2$ \Comment{Update target networks } 
	\EndFor
\EndFor
\Ensure $\theta_1$, $\theta_2$, $\phi$\Comment{Optimized parameters}
\end{algorithmic}
\end{algorithm}

\section{Hyper parameters}
\label{app:hyper_parameter}

Table~\ref{tab:shared_params} lists the common SAC parameters used in the comparative evaluation in Figure~\ref{fig:overall_result}.

\begin{table}[H]
\renewcommand{\arraystretch}{1.1}
\centering

\vspace{1mm}
\begin{tabular}{l l| l }
\toprule
\multicolumn{2}{l|}{Parameter} &  Value\\
\midrule
\multicolumn{2}{l|}{\it{Shared}}& \\
& optimizer &Adam \\
& learning rate & $3 \cdot 10^{-4}$\\
& discount ($\discount$) &  0.99\\
& replay buffer size & $10^6$\\
& number of hidden layers (all networks) & 2\\
& number of hidden units per layer & 256\\
& number of samples per minibatch & 256\\
& nonlinearity & ReLU\\
\midrule
\multicolumn{2}{l|}{\it{SAC}}& \\
& target smoothing coefficient ($\tau$)& 0.005\\
& target update interval & 1\\
& gradient steps & 1\\
\midrule
\multicolumn{2}{l|}{\it{OAC}}& \\
& \text{beta UB} & 4.66\\
& \text{delta} & 23.53\\
\midrule
\multicolumn{2}{l|}{\it{GAC}}& \\
& dynamic weight  ($\beta_t$)& epoch number*1\\
& sample range ($s_r$) & 7\\
& sample size ($s_n$)& 32\\
\bottomrule
\end{tabular}
\caption{GAC Hyper-parameters}
\label{tab:shared_params}
\end{table}

\section{Environment properties}
\label{app:environment_properties}
The properties of each environment are summarized in Table~\ref{table:environment_details}.

\begin{table*}[ht]
\begin{center}
\begin{tabular}{l|cccc}
\toprule
Environment & State dim & Action dim & Episode Length \\
\midrule
Humanoid-v2 & 376 & 17  & 1000 \\
Ant-v2 & 111 & 8  & 1000 \\
HalfCheetah-v2 & 17 & 6  & 1000 \\
Walker2d-v2 & 17 & 6  & 1000 \\
Hopper-v2 & 11 & 3  & 1000 \\
Swimmer-v2 & 8 & 2  & 1000 \\

\bottomrule
\end{tabular}
\end{center}
\caption{The details of Mujoco Environments used in this paper.} 
\label{table:environment_details}
\end{table*}
\section{More results}\label{sec:more_results}

\subsection{Relation between beta and sample size}
We sample $n$ integers uniformly from [1,1000] and give the numerical result of the $\text{lse}_{\beta}({\bf{X}})$, $\text{sm}_{\beta}({\bf{X}})$ and $\frac{1}{\beta}H({\bf{X}})$. According to the results shown in Table~\ref{table:fix_beta_0.01},Table~\ref{table:fix_beta_1},Table~\ref{table:fix_beta_100}, we can see that a large beta will lead $\text{sm}_{\beta}({\bf{X}})$ give a consistent result with the max operator. And, $\text{sm}_{\beta}({\bf{X}})$ is better than $\text{lse}_{\beta}({\bf{X}})$, it approximate the maximum from the lower bound. With a large beta, the maximum value can be aprroximated regardless of the sample size.

\begin{table*}[ht]
\begin{center}
\begin{tabular}{l|ccccc}
\toprule
$n$, $beta$=0.01 & $\text{lse}_{\beta}({\bf{X}})$ & $\text{sm}_{\beta}({\bf{X}})$ & $\frac{1}{\beta}H({\bf{X}})$ & maximum \\
\midrule
10 & 992.46 & 932.61  & 59.85 & 976 \\
100 & 1252.17 & 900.62  & 351.50 & 995 \\
1000 & 1461.84 & 899.60  & 561.094 &999 \\
10000 & 1692.55 & 900.98  & 779.09 & 999 \\
100000 & 1920.21 & 899.25  & 896.54 & 999 \\
1000000 & 2150.74 & 899.56  & nan &999 \\
\bottomrule
\end{tabular}
\end{center}
\caption{The results when $beta$=0.01.} 
\label{table:fix_beta_0.01}
\end{table*}

\begin{table*}[ht]
\begin{center}
\begin{tabular}{l|ccccc}
\toprule
$n$, $beta$=1 & $\text{lse}_{\beta}({\bf{X}})$ & $\text{sm}_{\beta}({\bf{X}})$ & $\frac{1}{\beta}H({\bf{X}})$ & maximum \\
\midrule
10 & 834.69 & 834.00  & 0.69 & 834 \\
100 & 970.02 & 969.92  & 0.09 & 970 \\
1000 & 1000.29 & 998.70  & 1.58 &999 \\
10000 & 1001.69 & 998.31  & 3.38 & 999 \\
100000 & 1004.10 & 998.41  & 5.68 & 999 \\
1000000 & 1006.36 & 998.41  & 7.85 &999 \\
\bottomrule
\end{tabular}
\end{center}
\caption{The results when $beta$=1.} 
\label{table:fix_beta_1}
\end{table*}

\begin{table*}[ht]
\begin{center}
\begin{tabular}{l|ccccc}
\toprule
$n$, $beta$=100 & $\text{lse}_{\beta}({\bf{X}})$ & $\text{sm}_{\beta}({\bf{X}})$ & $\frac{1}{\beta}H({\bf{X}})$ & maximum \\
\midrule
10 & 947.0 & 947.0  & 0.0 & 947 \\
100 & 996.0 & 996.0  & 0.0 & 996 \\
1000 & 997.0 & 997.0  & 0.0 &997 \\
10000 & 999.02 & 999.00  & 0.02 & 999 \\
100000 & 999.05 & 999.0  & 0.05 & 999 \\
1000000 & 999.07 & 999.0  & 0.07 &999 \\
\bottomrule
\end{tabular}
\end{center}
\caption{The results when $beta$=100.} 
\label{table:fix_beta_100}
\end{table*}

\subsection{Reward difference between exploration and evaluation}
\begin{figure}[ht]
    \centering
    \includegraphics[width=0.98\linewidth]{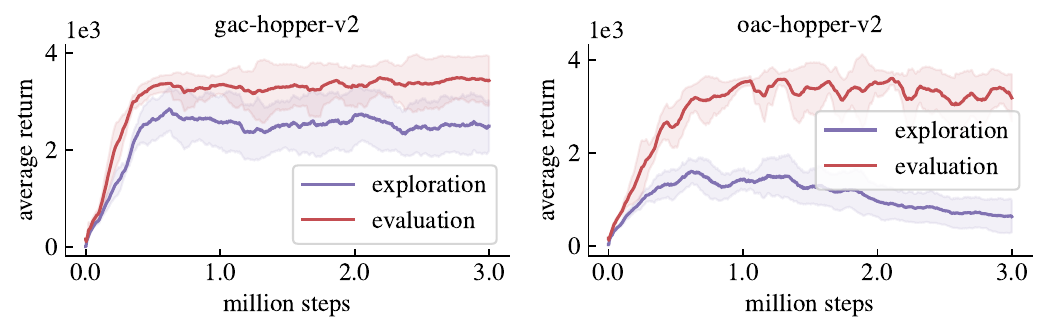}
    \caption{Exploration and evaluation difference of GAC and OAC}
    \label{fig:reward_difference}
\end{figure}

As shown in Figure~\ref{fig:reward_difference}, Because both algorithms are related to exploration, the evaluation return is higher than the exploration return. However, something goes wrong in OAC exploration. As we use transition sampled from the replay buffer to train the policy, it does not seem to have much impact on policy learning. Instead, it shows that our exploration strategy is better than the OAC method.

\subsection{Reward comparison between GAC and OAC}
\begin{figure}[ht]
    \centering
    \includegraphics[width=0.98\linewidth]{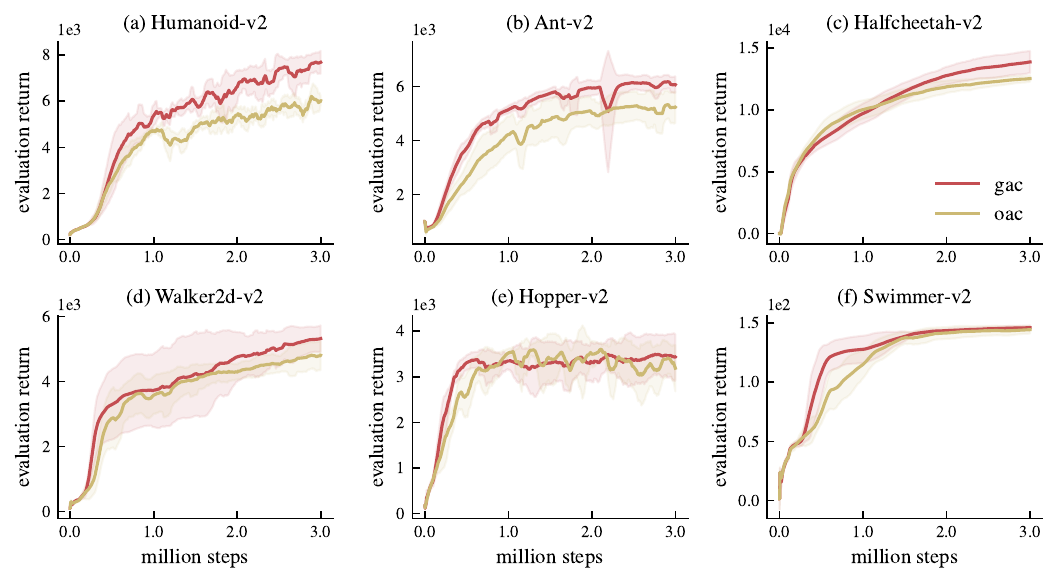}
    \caption{Average evaluation episode return of GAC and OAC}
    \label{fig:eval_gac_oac}
\end{figure}

\begin{figure}[ht]
    \centering
    \includegraphics[width=0.98\linewidth]{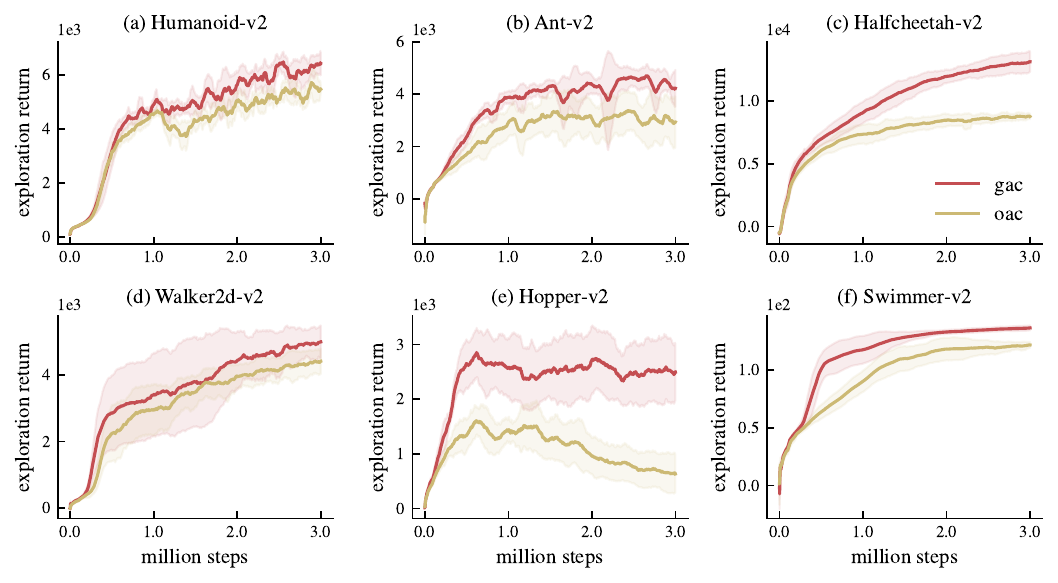}
    \caption{Average exploration episode return of GAC and OAC}
    \label{fig:exp_gac_oac}
\end{figure}

As shown in Figure~\ref{fig:eval_gac_oac}, our method performs better in policy evaluation. Another thing to note is that our method is inspired by OAC, and we found some problems with the exploration strategy of OAC. Therefore, we need to prove that our exploration strategy is better, as shown in Figure~\ref{fig:exp_gac_oac}, this figure shows that our exploration strategy is better.

\subsection{More hyper-parameter experiments}
\begin{figure}[ht]
    \centering
    \includegraphics[width=0.98\linewidth]{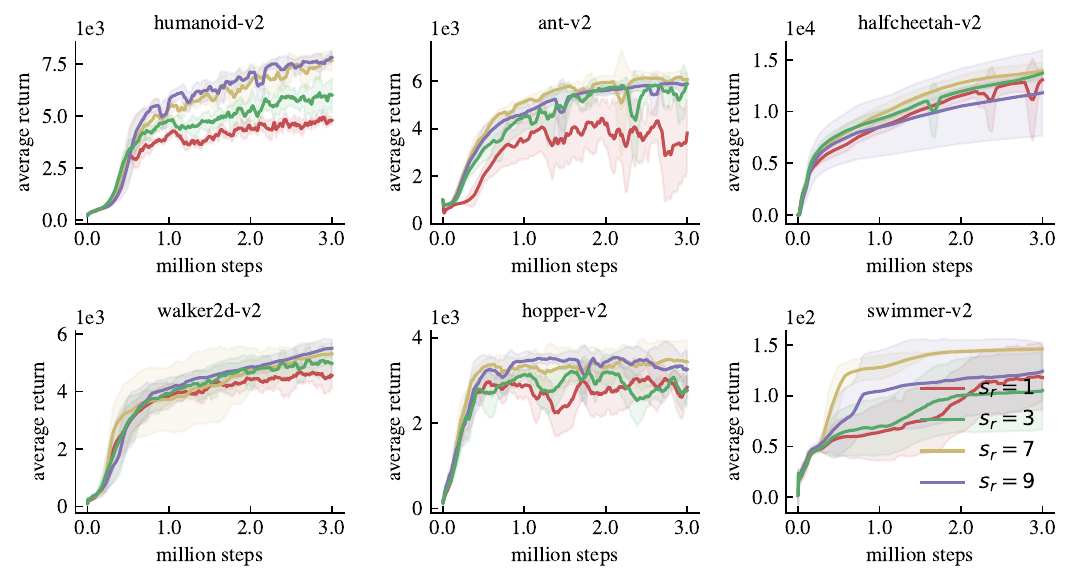}
    \caption{The results of different sample range on the six environments. }
    \label{fig:full_range}
\end{figure}

\begin{figure}[ht]
    \centering
    \includegraphics[width=0.98\linewidth]{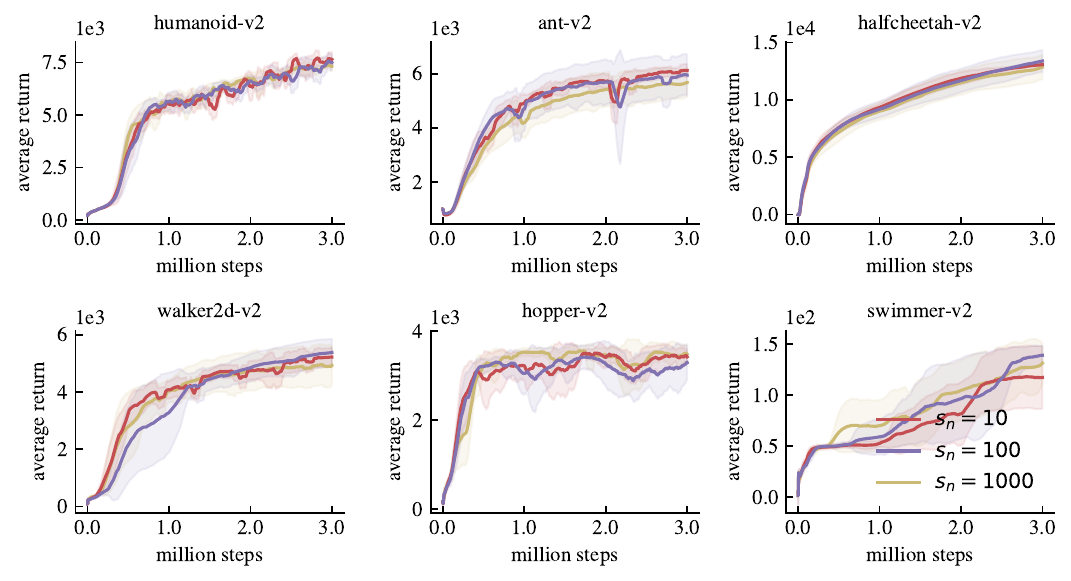}
    \caption{The results of different sample size on the six environments. }
    \label{fig:full_size}
\end{figure}

\begin{figure}[ht]
    \centering
    \includegraphics[width=0.98\linewidth]{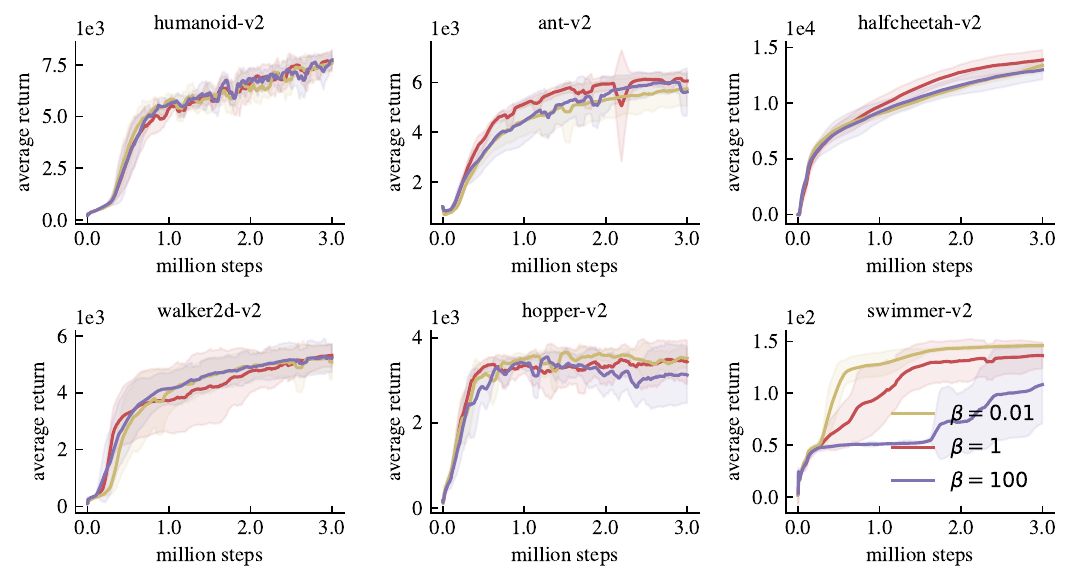}
    \caption{The results of different beta on the six environments. }
    \label{fig:full_beta}
\end{figure}

We show all the results of different hyper-parameters on these six environments. As shown in Figure~\ref{fig:full_range}, Figure~\ref{fig:full_size} and Figure~\ref{fig:full_beta}, combine the numerical results, we can better choose the value for hyper-parameters.

\subsection{More visualization for the Q value}
\begin{figure}[htbp]
    \centering

    \subfigure[Epoch 0]{
        \includegraphics[width=0.48\textwidth, trim={25mm 0mm 20mm 20mm}, clip]{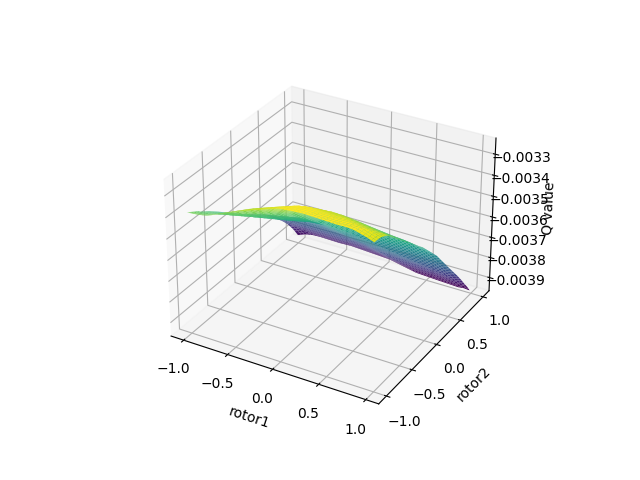}
    }
    \subfigure[Epoch 500]{
        \includegraphics[width=0.48\textwidth,  trim={25mm 0mm 20mm 20mm}, clip]{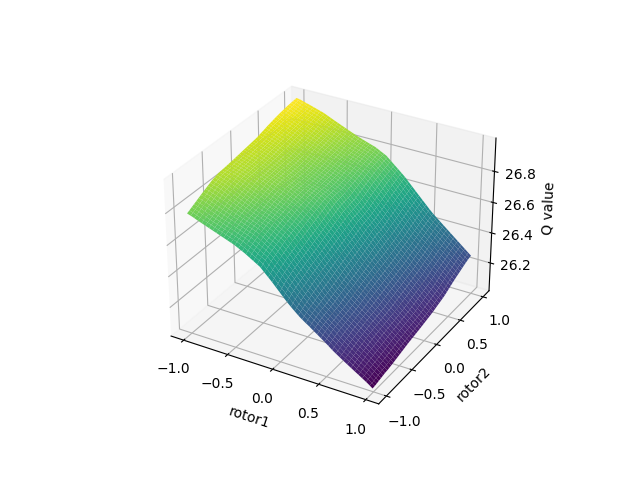}
    }
    \subfigure[Epoch 1500]{
        \includegraphics[width=0.48\textwidth, trim={25mm 0mm 20mm 20mm}, clip]{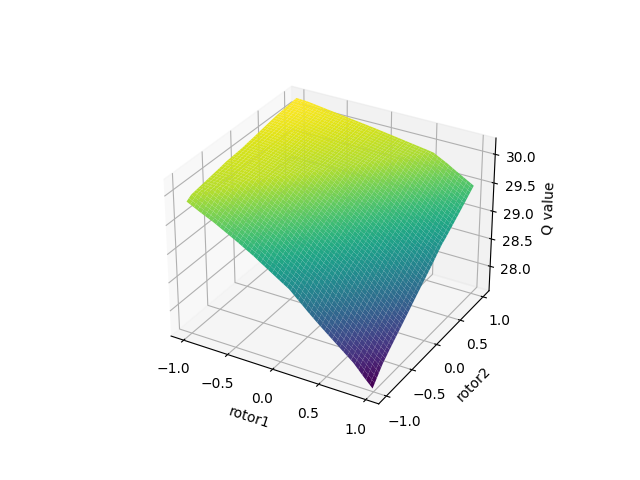}
    }
        \subfigure[Epoch 2000]{
        \includegraphics[width=0.48\textwidth, trim={25mm 0mm 20mm 20mm}, clip]{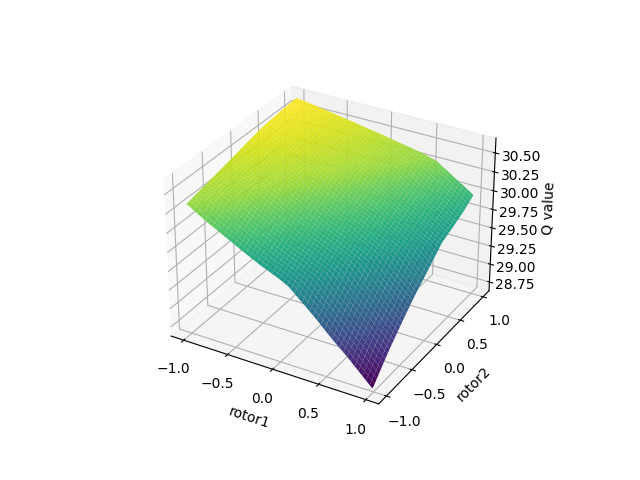}
    }

    \caption{3D surfaces of different epoch Q function} 
    \label{fig:visulization_Q_2}
\end{figure}

To observe the surface of the Q network, we plot different stage Q value, which is evaluated based on a random start state and sampled actions in the Swimmer-v2 environments. As shown in Figure~\ref{fig:visulization_Q_2},in the initial stage, the surface is not flat, which is influenced by the input (state and action); if the state and action is zero vector, this surface should be flat, all zero. This phenomenon shows that neural networks imply prior knowledge about choosing actions. Policy initialization is closely related to policy learning. As training progresses, the final optimal action dramatically differs from the initial policy.

\section{Limitations and broader impacts} 
\paragraph{Limitations} In low-dimensional action spaces, our method shows little improvement. It is particularly noticeable that in the swimmer-v2 environment, state-of-the-art results can reach an episode reward of 350. Furthermore, exploration costs should also increase as the action space's dimensionality increases. The conclusions we drew earlier may have limitations. However, it is challenging to develop environments with higher-dimensional action spaces, and we still need to fully validate our conclusions in such environments.

\paragraph{Broader impacts} We do not anticipate any negative consequences from using our method in practice.

\clearpage
\newpage
\begin{figure}[ht]
    \centering

    \subfigure[HumanoidFlagrun-v1]{
        \includegraphics[width=0.42\textwidth, trim={0mm 2mm 0mm 2mm}, clip]{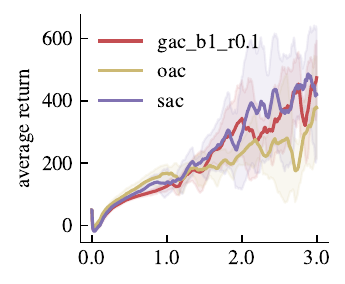}
    }
     \hfill
    \subfigure[HumanoidFlagrunHarder-v1]{
        \includegraphics[width=0.42\textwidth,  trim={0mm 2mm 0mm 2mm}, clip]{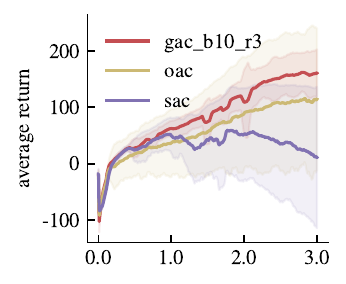}
    }

    \caption{Additional evaluation in the Roboschool simulation. We use different hyper-parameters. ``gac\_b1\_r0.1'' means that $\beta=1$ and $s_r=0.1$.} 

\end{figure}

\begin{figure}[h]

\centering
    \subfigure[random vs greedy]{
        \centering
        \includegraphics[width=0.42\textwidth, trim={0mm 2mm 0mm 2mm}, clip]{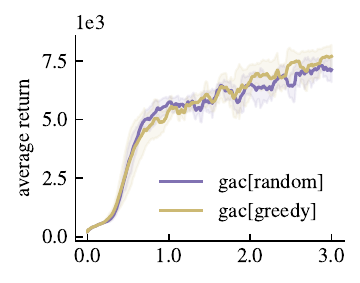}
    }
    \hfill
    \subfigure[large range effect]{
        \centering
        \includegraphics[width=0.42\textwidth,  trim={0mm 2mm 0mm 2mm}, clip]{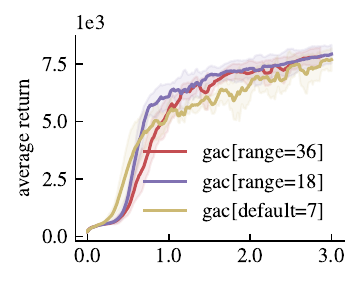}
    }
    \subfigure[stable learning and greedy exploration]{
        \centering
        \includegraphics[width=0.42\textwidth, trim={0mm 2mm 0mm 2mm}, clip]{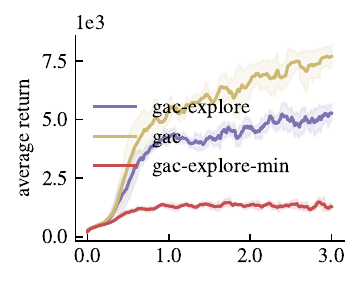}
    }
    \hfill
        \subfigure[more Q network]{
            \centering
        \includegraphics[width=0.42\textwidth, trim={0mm 2mm 0mm 2mm}, clip]{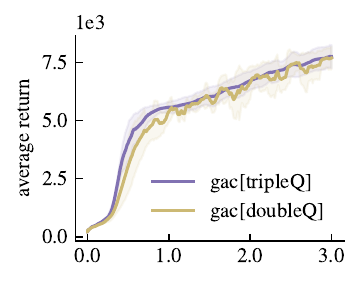}
    }
    \caption{More results in the Humanoid-v2 environment. (a) ``random'' means randomly taking $Q_1(s, a)$ or $Q2(s, a)$ when constructing $\pi_E$. (b) ``range\=36'' means $s_r$=36. (c) ``gac-explore'' means expore without greedy Q, just with policy. Furthermore, ``gac-explore-min'' means no use of the min value as the target value. (d) ``tripleQ'' means the value function ensemble framework comprises triple Q networks.} 

\end{figure}

\end{document}